# What your brain activity says about you:

# A review of neuropsychiatric disorders identified in resting-state and sleep EEG data

**Authors:** Scanlon, J.E.M.[1], Pelzer, A. [1], Gharleghi, M. [2], Fuhrmeister, K.C. [2], Köllmer, T.[2], Aichroth, P. [2], Göder, R.[4], Hansen, C.[3], Wolf, K.I. [1]

**Affiliations**

[1] Fraunhofer Institute for Digital Media Technology, Branch Hearing, Speech and Audio Technology, Oldenburg, Germany

[2] Fraunhofer Institute for Digital Media Technology, Ilmenau, Germany

[3] Department of Neurology, University Medical Center Schleswig-Holstein (UKSH), Kiel, Germany

[4] Department of Psychiatry and Psychotherapy, University Medical Center Schleswig-Holstein (UKSH), Kiel, Germany

**Corresponding author:**

Dr. Joanna E. M. Scanlon

Group Mobile Neurotechnologies (MNT)

Fraunhofer IDMT, Branch Hearing, Speech and Audio Technology HSA

Marie-Curie-Strasse 2, Oldenburg, Germany

joanna.scanlon@idmt.fraunhofer.de

**Abstract:**

**Introduction:** Electroencephalogram (EEG) monitoring devices and online data repositories hold large amounts of data from individuals participating in research and medical studies without direct reference to personal identifiers. However, beyond the intended uses of these methods, machine learning studies have shown the ability to re-identify persons and detect personal health information including mental and sleep disorders from resting state and sleep EEG. The combination of re-identifiability and the presence of sensitive information presents a particular challenge in the context of data protection. This paper explores what types of personal and health information have been detected and classified within task-free EEG data. Additionally, we investigate key characteristics of the collected resting-state and sleep data, in order to determine the privacy risks involved with openly available EEG data. **Methods:** We used Google Scholar, Web of Science and searched relevant journals to find studies which classified or detected the presence of various disorders and personal information in resting state and sleep EEG. Only English full-text peer-reviewed journal articles or conference papers about classifying the presence of medical disorders between individuals were included. A quality analysis carried out by 3 reviewers determined general paper quality based on specified evaluation criteria. **Results:** In resting state EEG, various disorders including Autism Spectrum Disorder, Parkinson's disease, alcohol use disorder have been classified with high classification accuracy (>90%), often requiring only 5 mins of data or less. Sleep EEG data tends to hold classifiable information about sleep disorders such as sleep apnea, insomnia, and REM sleep disorder, but usually involve longer recordings or data from multiple sleep stages. **Discussion:** Many classification methods are still developing but even today, access to a person's EEG can reveal sensitive personal health information. With an increasing ability of machine learning methods to re-identify individuals from their EEG data, this review demonstrates the importance of anonymization, and the development of improved tools for keeping study participants and medical EEG users' privacy safe.

## Introduction

Open data sharing has become more popular recently, with many open neuroscience repositories now available for sharing EEG data (e.g. openneuro.org), as well as special data standards such as BIDS for more efficient data sharing (Jeung et al., 2024). Additionally, EEG data is often shared in repositories or similar systems when patients use Brain computer interfaces (BCIs; Xia et al., 2023; Li et al., 2015), as well as home sleep (e.g. Guillot et al., 2020) or long-term monitoring systems, e.g. in the context of epileptic seizure detection (Wong et al., 2023). These repositories allow us to share data between clinics and researchers for analysis, and even share data with other researchers for re-analysis to ensure reproducibility in science and to publish new results (e.g. Azim et al., 2010; Ziogas et al., 2023). This can lead to many benefits to the scientific world, which include improving research efficiency, democratizing knowledge, and fostering the development of novel analysis approaches (Arza & Fressoli, 2017).

A key requirement for putting EEG data into an online repository is that there is no direct identifying information attached to it, such as individual participants' names. This is known as anonymization, which is defined as the removal of all personalized and personal information such as an individual's name and age from the research data (Meyermann & Porzelt). According to the General Data Protection Regulation (GDPR), which imposes data privacy and security laws onto any organizations which target or collect data related to people in the European Union, personal data must have identifying information removed as soon as it is no longer necessary for the data collectors purposes (Art. 5: Principles relating to processing of personal data). Any personal data should be processed in a way that ensures appropriate security. Additionally, sensitive personal data such as ethnicity, religion, political opinion, union membership, philosophical beliefs, sexuality and health are given special protection, and may only be used under special circumstances (GDPR Art. 9: Processing of special categories of personal data). These laws exist in order to protect the privacy of individuals who participate in research studies.

But what if EEG data donors could be identified based on features within the data, without a name being attached? Studies have shown some ability to use machine learning to re-identify individuals, simply using their resting state (Zhang et al., 2018a; Zhang et al., 2020; Das et al., 2019; for a review, see Del Pozo-Banos et al., 2014) or sleep EEG data (De Gennaro et al., 2005). If a direct personal identifier is never made accessible, this would not be critical. But if just one data set reveals the identification directly or based on meta-information, other data sets could be de-anonymised by the mentioned methods. The possibility of a personal identifier related to an EEG dataset getting published in the future therefore poses a risk for the data donor if sensitive information in the EEG has not been removed.

The risk of reidentification is a growing concern for various types of data, including computer vision (Gong et al., 2014), wearable photoplethysmography, electrodermal and accelerometer data (Alam, 2021; Sullivan & Alam, 2022), as well as data using various combinations of ECG, EMG, and EEG (for a review, see Chikwetu et al., 2022) including BCI data (for a review, see Xia et al., 2022). Indeed, EEG data have shown enough interpersonal variation that they have even been suggested for use for authentication systems (Jayarathne et al., 2016; Klonovs et al., 2012; Shedeed, 2011; for a review, see Campisi et al., 2012 and Chan et al., 2018). Reidentification is an important ethical concern for clinical brain data, as it allows the breach of patient or research participant privacy (Kellmeyer, 2021). With growing concerns of reidentification of neuroscientific data, some studies suggest legislation to go beyond anonymization into harm reduction, to prevent the use of this data in decisions about health insurance and employment (Jwa et al., 2022).

A recent survey on methods in the field of EEG based authentication found that resting states (either eyes open (REO) or eyes closed (REC)) were the most popular protocol for collecting EEG signals for the purpose of authentication (Bidgoly et al., 2020). Task-free EEG data such as REO, REC and sleep data could hold particularly important information about individuals, because they have both been shown to hold biometric information about an individual (De Gennaro et al., 2005; Chikwetu et al., 2022; Del Pozo-Banos et al., 2014). Additionally, in contrast to EEG authentication methods which tend to use event-related potentials (ERP) as brain responses to specific sensory or motor events, resting state and sleep EEG do not require response to a task and are therefore easier to reproduce (Bidgoly et al., 2020).

If individuals are identifiable by their resting state EEG data, what additional information can be found in this data? Various data features in resting state and sleep EEG data can point to mental disorders such as sleep disorders (Azim et al., 2010), depression (Shim et al., 2023), as well as addictions such as smoking (Rass et al., 2016) and alcoholism (Mumtaz and colleagues, 2017). For example, a study by Mumtaz and colleagues (2018) was able to obtain a classification accuracy of 98% when classifying individuals with alcohol use disorder in comparison to age-matched controls. Zhao et al. (2017) found significant alpha band differences in 5 minutes of resting state EEG data between controls and individuals who were addicted to heroin but abstaining at the time of data collection. Another study by Ziogas and colleagues (2023) used EEG data to distinguish between homosexual and heterosexual males at a rate up to 83%. The aforementioned study examples were not all included in this particular review, but indicate examples of private information attempted to be distinguished through EEG. The current review intends to look deeper into the information available in resting state and sleep EEG data.

This work is a comprehensive review of the available studies on what type of information about individuals can be found in task free (e.g. sleep data, REO, REC) EEG data. In particular, we aim to answer the following questions:

- What personal health information can be detected using machine learning in task-free EEG data?
- How does sleep data compare to resting state EEG data (e.g. number of subjects, recorded time, number of channels) when it comes to classifying medical information?
- Does the quality and reproducibility of these papers indicate that identifying medical disorders in publicly available EEG data is a significant threat?

We focus on studies that compare the extraction of certain personal health information from EEG data during sleep and during resting state. We also compare the type of data used (e.g. REC or REO; sleep staged or not) and amounts of data available (e.g. recorded signal time, sample size, number of channels), as well as main classification results. The goal of this paper is to provide an overview of the sensitive personal health information contained in task-free EEG data that can be revealed with contemporary machine learning techniques.

## Research methodology

The search for papers was executed in two phases. First, a preliminary search in google scholar was carried out in order to establish a scientific base of the field, with terms including ‘EEG’, ‘disorder’, ‘classification’, ‘detection’, and ‘personal information’, until December 15, 2023. Second, a more structured search was carried out through a Web of Science search using the following keyword terms: ‘detect*’, ‘disorder’, ‘EEG’, ‘classif*’, ‘sleep’, and ‘resting-state’, combined in various ways, until January 4, 2024. Studies found in these two searches were entered into Covidence, (Covidence, Australia). Additionally, we searched through reference sections of relevant papers, until March 26, 2024. All studies were reviewed by one reviewer and were required to be full-text peer-reviewed journal articles or conference papers in English published before January 1, 2024. They had to focus on either classifying the presence of a medical disorder from the data of individual persons. Papers were excluded if they required more than resting state EEG or sleep EEG/polysomnography (PSG) data to classify targets. Additionally, we excluded papers which had similar findings to included studies by the same authors, (similar to Chikwetu et al., 2022), as well as any papers which were not fully accessible from our libraries, or lacked clarity or important information. The exclusion process is detailed in the PRISMA chart in Figure 1 (Page et al., 2021).

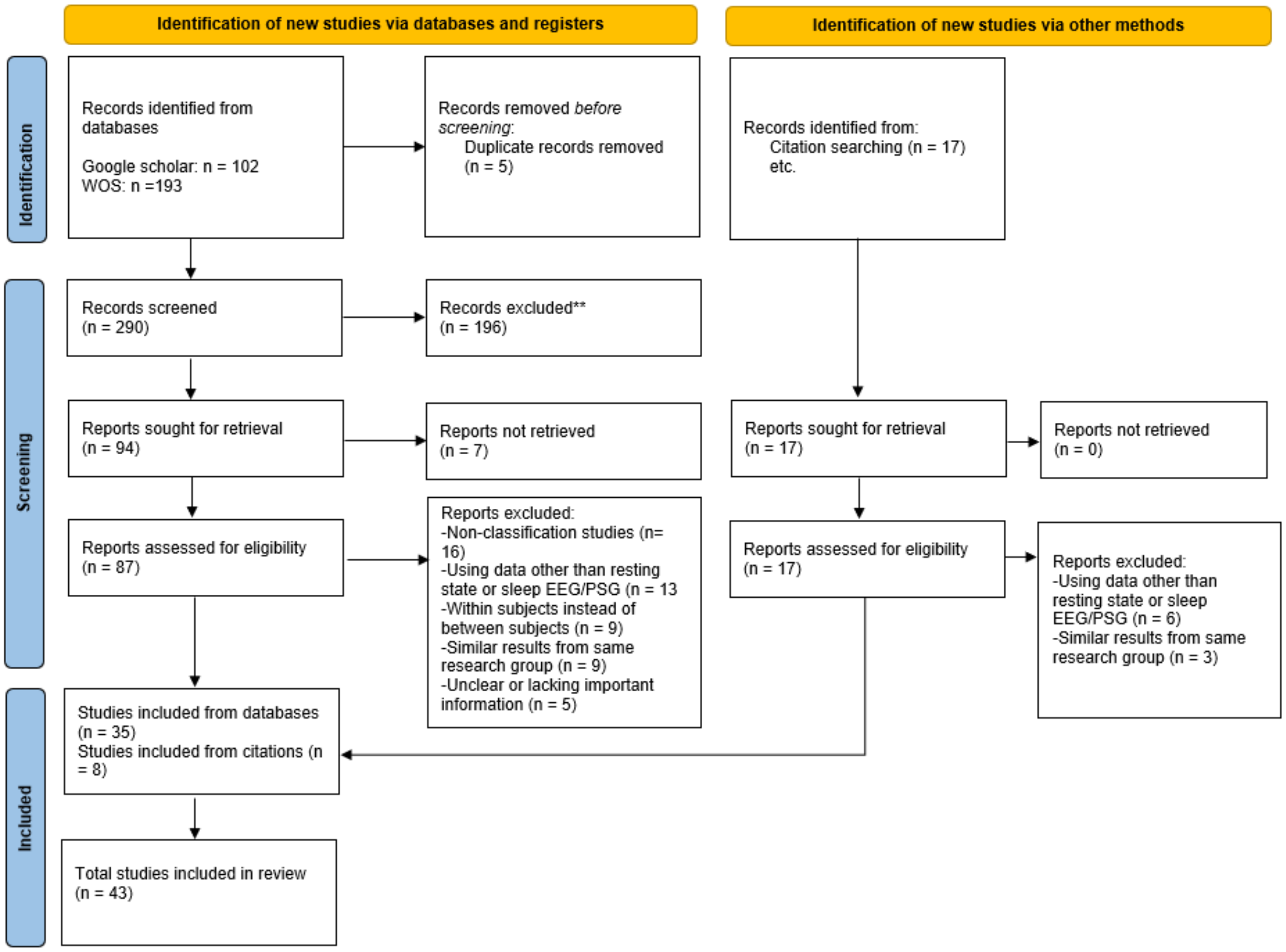

**Figure 1.** PRISMA flow chart of papers collected and screened to be included in the review.

*Quality Analysis*

To assess quality of our included studies, we used a custom quality assessment tool based on Chikwetu et al., 2022; see Supplementary Material). In addition to questions about the validity of the study, our questions assessed quality of the machine learning procedures, and reproducibility of the methods (e.g. code and data availability). For this we had three reviewers (J.E.M.S., A.P., and M.G.) go over all papers. Each paper was reviewed by at least two of the reviewers while the third adjudicated any disagreements.

*Data extraction and Synthesis*

From the included studies, we extracted 8 study characteristics to assess the type of data required to classify disorders from resting state EEG data, and the quality of this classification. These were: sample size, number of channels used, amount of EEG signal time required for classification (training and testing, before artifact correction), best performing classifier and features used, and best classification results (e.g. classification accuracy or sensitivity, specificity, etc.). Plots and computations were performed using Python 3, as well as external libraries including NumPy (Harris et al., 2020), Pandas (McKinney, 2011) and Matplotlib (Ari, & Ustazhanov, 2014; Hunter et al., 2007).

## Results

Our searches retrieved 295 studies altogether, resulting in 290 studies to be screened after removing 5 duplicates (Figure 1). After title and abstract screening, 196 studies were excluded, leading to 94 studies advancing to the full-text review. Of these, 7 were not available in full-text form, and 52 were found to not meet our eligibility criteria, leaving 35 studies. Additionally, searching through the reference sections of relevant papers lead to finding 17 more studies. Upon assessing these studies for eligibility, 8 were found to be eligible for this review. Altogether, 43 studies were included in this review.

Of the studies included, 26 used resting state EEG data, and 17 used sleep EEG data or PSG to classify disorders between individuals. One study was considered as a sleep study because it was centered around daytime excessive sleepiness and used sleep-staged EEG data (Melia et al., 2015), but used only data classified as waking state, which could be considered very similar to resting state data. Because of this ambiguity, we did not consider this study in either the resting state or sleep EEG data group when computing averages (e.g. in recording time, channel numbers, classification accuracy). Two sleep studies included 2 datasets which fit our criteria, and therefore each of these datasets were counted as a single study when counting averages for sample size, channel number and recording time, and classification accuracy.

In the included studies, 36 unique disorders were classified (see Table 1 & 2 and Figure 2). Of these, 24 disorders were primarily identified with resting state EEG data, with the most common being Major Depressive Disorder (MDD) or depressive disorders (5 studies). Sleep data was primarily used to identify 12 disorders, with the most common being insomnia (10 studies). Most studies classified individuals with one disorder compared to normal controls, but 5 studies classified between 2 or more disordered groups and normal controls. Only one study did not include a healthy/normal control group (Kim et al., 2023).

*Quality assessment:*

The resting state studies included were of generally high quality, according to our custom quality assessments (see Supplementary Material). Only one study was suspected of possible publication bias, due to a list of competing interests and funding sources directly related to the disorder classified (Garcés et al., 2022). Three (11.5%) resting state studies were considered of moderate quality, due to missing information assessed in our quality assessment tool, while the rest (23; 88%) were considered high quality. The sleep studies on the other hand were of lower quality on average, with 7 (41%) found to be high quality, 9 (35%) found to be moderate quality, and 1 of low quality, according to our assessment.

[Tables 1, 2, & 3 here. Tables included in separate document]

*Sample size and channel numbers*

Resting state studies tended to use more subjects, with an average of 229 participants per dataset while sleep studies used about 40 participants per dataset (Figure 2). The average number of subjects in resting state datasets was greatly increased by one outlier, in which Langer et al., (2022) had 1921 subjects. However, with this study removed, the average for resting state studies is 161.

Similarly, resting state studies also tended to use more channels for measurements, with an average of 34.3 channels, while sleep studies used on average of 3.4 channels per dataset (two studies included 2 eligible datasets with different numbers of channels). Six sleep studies (35.3%; Sharma et al., 2023; Qu et al., 2021; Yang & Liu, 2020; Hamida et al., 2016; Kuo and Chen, 2020; Wadichar et al., 2023) only used primarily one channel, while only two (7.7%; Liechti et al., 2013; Sharma & Acharya, 2021) resting state studies used only one channel in their analysis. Two more resting state studies used single channels in a subset (Ghoreishi et al., 2021; Suuronen et al., 2023), but these both achieved better results with more channels.

*Signal time*

In Tables 1 and 2, the main study characteristics are listed. In the resting-state EEG studies, relatively low amounts of signal time were required to detect disorders (Figure 3). The minimum amount of recording time used was 30 s (Suuronen et al., 2023), with the maximum being 15 min (mean = 4.7 min). For sleep data, the amount of signal time available for data testing and training was sometimes unclear. Two studies did not clearly specify the number of recordings or amount of time required for the analysis for each subject (Gholami et al., 2022; Wadichar et al., 2023). Three studies specified that data was used from one night's sleep (Hamida et al., 2016; Shahin et al., 2017; Vishwanath et al., 2021). 53% of sleep studies specified either a range or number of epochs or amount of time used in the analysis for each participant (Aydin et al., 2015; Kuo and Chen, 2020; Moussa et al., 2022; Sharma et al., 2023; Qu et al., 2021; Heyat et al., 2019; Hansen et al., 2013; Yang & Liu, 2020; De Dea et al., 2019), while others gave an approximate recording time. To compute an approximate average recording time over all sleep studies, we used the minimum recording time listed for each study where this information was available. In studies that specified that one night's sleep was used, we counted 8 hours, as the standard recommended sleep time for an adult is 7 hours or more , and this would include time for falling asleep and waking (Watson et al., 2015). The average amount of recording time for sleep studies was 5.3 hours. The longest recording of sleep used for a study was 9 hours (Malinowska et al., 2013; Sharma et al., 2023) while the shortest was < 30 minutes (Kuo & Chen, 2020).

*Classification accuracy comparison*

91% of the studies included used classification accuracy as their main dependent variable, therefore we used this measure when comparing study outcomes. Generally, classification accuracy is defined as the ratio of the number of correct predictions to the total number of predictions. To compare classification accuracy between the two data types, we took the best classification accuracies for each study that used classification accuracy as a measure and averaged them over resting state and sleep studies. Resting state had a mean of 85.7 % accuracy while sleep data had a mean of 90.3% accuracy.

*Type of data used*

In the resting state studies, the most common form of resting state used was with eyes closed, with 15 studies (57.7%) using only REC data, and six (23%) using both REO and REC data. Three studies (11.5%) used only REO data, while two did not state specifically whether the participants were asked to open or close their eyes during data recording.

For the sleep studies, 14 studies (83.3%) used sleep staged EEG data for their classification in some way. Two studies did not separate sleep data by stages (Gholami et al., 2022; Malinowska et al., 2013) and one study specifically observed data before, during and after sleep spindles (de Dea et al., 2019).

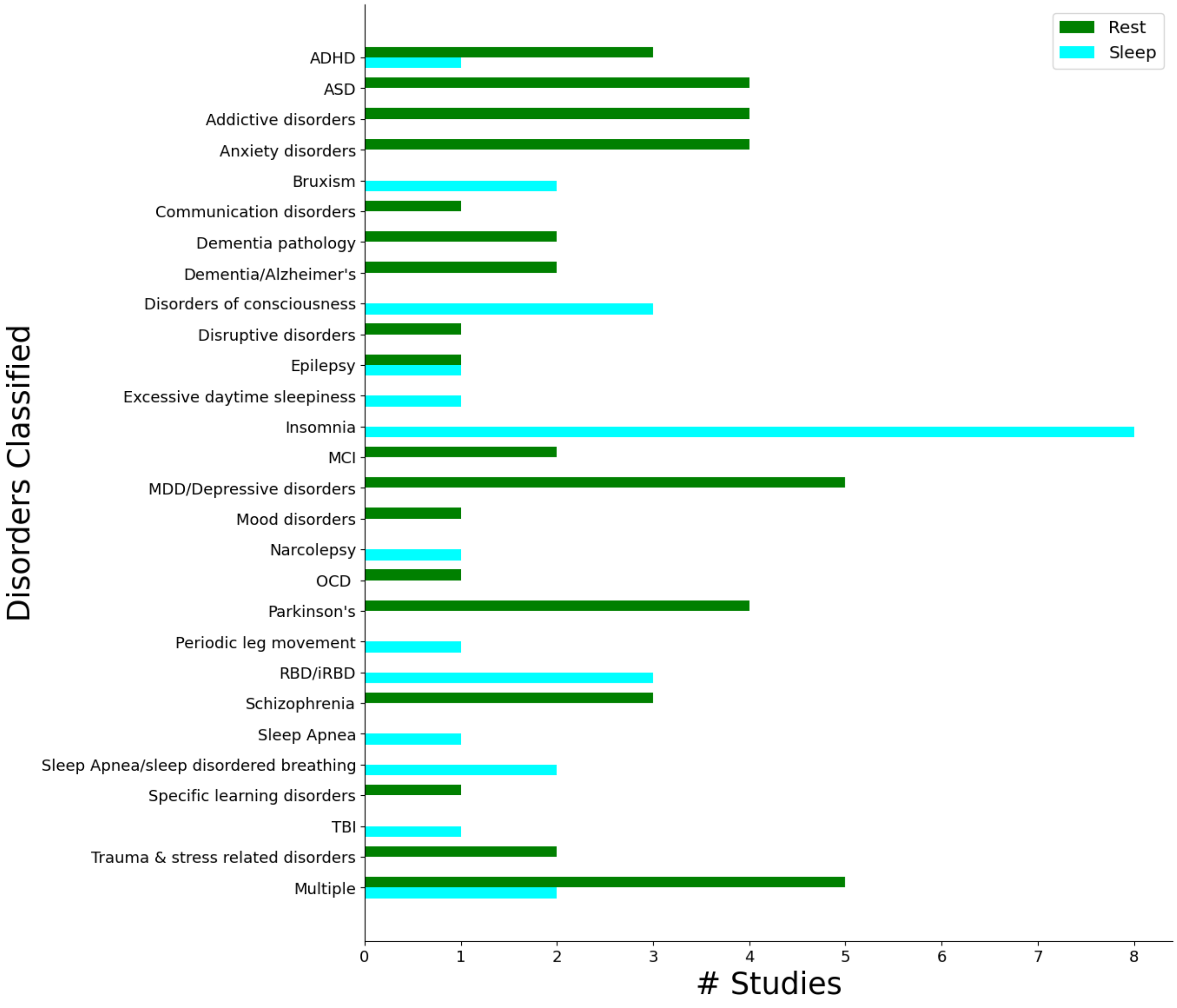

**Figure 2.** Proportions of types of disorders classified in resting state and sleep data. Some specific disorders are grouped together for plotting purposes. For papers classifying multiple disorders, each disorder is counted once, and also counted with the label 'Multiple'.

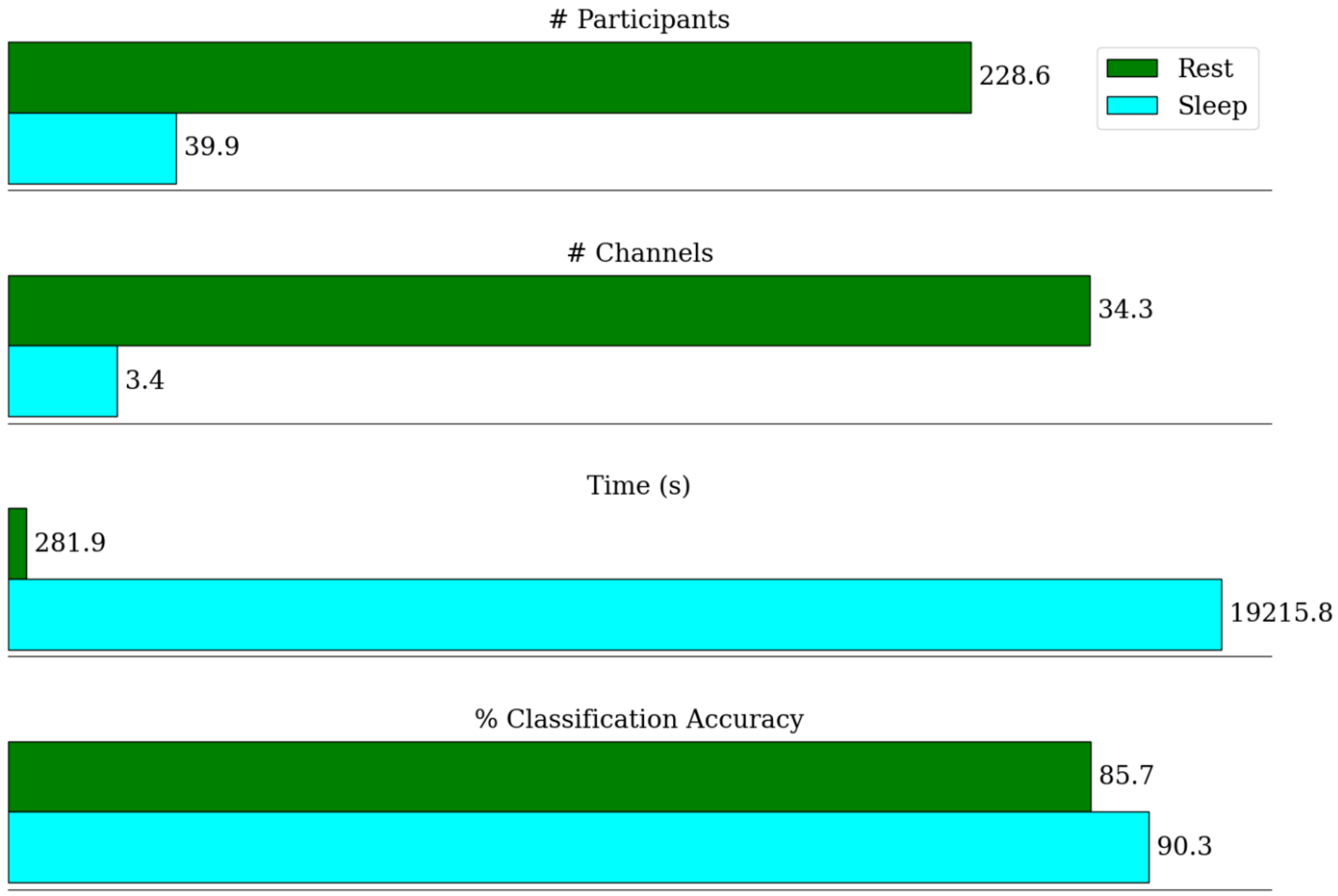

**Figure 3.** Mean number of subjects, channels, signal time length and classification accuracy for all studies included, in which this data was available.

## Discussion

This paper reviewed a variety of studies which used machine learning to detect medical disorders in resting state and sleep EEG data. In general, we can see that resting state data used more channels and subjects and less signal time for classifying disorders, while sleep data could often use fewer channels but required more signal time to get various sleep phases in the data. Average classification was generally high, but the quality analysis revealed that many studies lacked critical information.

*Quality analysis*

We originally aimed to select papers of only high quality, however the quality analysis showed that a substantial proportion of the studies used were of moderate or low quality, according to our assessment (see Supplementary Material). This was particularly true for the sleep EEG studies. These results are unsurprising, as it has been documented that machine learning papers are sometimes lacking in quality (Lipton & Steinhardt, 2019; Yusuf et al., 2019). Lipton & Steinhardt (2019) identified several writing trends in machine learning papers which have the potential of misleading readers. These include papers which are unable to distinguish between speculation and explanation, failure to correctly identify sources of empirical gains (e.g. Neural architecture vs. hyperparameter tuning), 'Mathiness', or the use of technical and complicated terms to obfuscate or impress the reader, as well as general misuse of language including suggestive word definitions, suitcase words, or words with many meanings, and overloaded terminology, or using words with specific meaning in an unspecific or contradictory way. Additionally, Yusuf et al. (2019) have reported that papers which use machine learning models specifically for medical diagnosis often fail to meet guidelines on reporting, for example, lacking adequate detail concerning the participant demographics, or details required for replication. This was the case for many papers featured here, as they often seemed to lack details that the authors and quality reviewers were looking for (e.g. details about machine learning and recording time) in order to evaluate them.

In particular, 8 (31%) resting state studies and 6 (35%) of the sleep studies recieved a 'somewhat' or 'no' in question 7 about quality of the machine learning performed in the study. For studies to receive a 'yes' in question 7 (see Supplementary material), they would have to demonstrate a good understanding of machine learning protocols by including information about the length of the EEG epochs used for training the machine learning model, as well as information about the way the data of individual participants is split between the training and test set, and whether this is done appropriately. The epoch length is important for adequately describing the model training and allowing replication of studies. Data splitting is important because it should be done without subject overlap between the training and test set (in most cases) to prevent unintended information leakage (Brookshire et al., 2024; Lee et al., 2023). As reidentification has been achieved with both resting state and sleep EEG data (Zhang et al., 2018a; Zhang et al., 2020; Das et al., 2019; Del Pozo-Banos et al., 2014; De Gennaro et al., 2005; Chikwetu et al., 2023), it's possible that individual subject traits within the EEG data can leak in both cases when attempting to detect psychiatric disorders (Brookshire et al., 2024; Lee et al., 2023). The model could then learn this additional information, unrelated to the classification labels, to boost its accuracy leading to an overestimation of a model's utility. It is seen as particularly problematic that the dataset may contain information that the data donor does not even know yet concerning potential early detection of diseases. Therefore, splitting without subject overlap also helps to prevent the unintended leakage of personal information.

*Resting state vs. Sleep data comparisons*

Sleep data studies often used substantially longer periods of signal time for their datasets than resting state EEG. This is partly due to the need for a full night's sleep in order to obtain enough data for all sleep stages, which requires at least one (often two, for habituation to the environment, e.g. Shahin et al., 2017) full night of sleep in a sleep lab. Additionally, this is one place where machine learning papers often lack reporting quality, as several papers were unclear with their reporting of the length of recording time, and consequently the exact amount of data used for training and testing the classifier. Some studies specify amounts of time within the sleep recording which was used for training and testing the classifier (e.g. Moussa et al., 2022; Qu et al., 2021). But other times papers only list the number of artifact-cleaned epochs that made it into the analysis, using vague amounts of time such as one night's sleep (Hamida et al., 2016; Shahin et al., 2017; Viswanath et al., 2021), or simply don't give a specific amount of time (Gholami et al., 2022). This can make it difficult for other studies to replicate their methods and findings. This complication in recording time reporting may have been more common in sleep papers due to the use of databases or the need to use longer recordings of somewhat noisy data, while in contrast resting state studies often could simply record relatively clean data for a short amount of time.

Sleep studies also had fewer participants than resting state studies. This is likely due to the large amount of effort required for sleep studies, as it would be much easier to recruit participants and carry out data collection for 15 minutes of data recording than one or two whole nights of sleep. With long and expensive data recordings, it is understandable that many sleep EEG papers chose to use already existing databases instead of collecting data (see Table 3).

Resting state studies used more channels, on average, to perform classification than sleep data studies (see Figure 3). Channels required for analysis likely depend on the specific disorder being detected. For example, Suuronen et al., (2023) compared varying channel numbers, as well as REC vs. REO data to detect Parkinson's disease, which is a neurodegenerative disease with several motor symptoms (Carlin et al., 2021). Suuronen et al. (2023) found that with larger channel numbers, the algorithm often selected central channels over the motor cortex. The authors also found that their best formation used REO data with 5 channels which were spread far away from each other, which the authors believe is because this allowed sampling from more independent EEG sources. In comparison, sleep studies often required only one or two channels because this is often enough data for adequate sleep staging (Lambert et al., 2023).

### *Limitations*

While many of the studies included in this review had high classification accuracies, it's hard to determine whether these disorders could actually be classified in an open dataset without any other information, as there are many differences between these datasets and a real world sample. Most of the studies included in this review used approximately equal numbers of individuals with disorders and healthy individuals, but in practical settings, the base rate of many disorders is much lower. For example, the WHO estimates that only 4.4% of the global population suffers from depressive disorder and 3.6% from an anxiety disorder (WHO, 2017), and any classifier attempting to detect these disorders would need to take this into account. Additionally, real world data will often include individuals with comorbidity, or individuals with multiple disorders at the same time, as well as varying factors such as age, which were controlled in most of these studies. Only one study in this collection dealt with comorbid disorders within individuals (sleep apnea and depression; Moussa et al., 2022), while most listed only one disorder within each participant. Additionally, most studies (where this information was available) age-matched their groups, while in real world data, age is often a factor in disorders such as Alzheimer's. Therefore, it is clear at the moment that detecting disorders blindly from public EEG data is not easily possible.

### *Conclusions*

This review demonstrates that task-free EEG data contains a large amount of information, including sensitive information about identity and health status. If an individual can be identified within a dataset through personal identifier or reidentification, this could put their dataset, as well as any related datasets at great risk for their privacy concerning identity and health information. At the moment the ability to detect this information in data without a direct personal identifier appears to be hindered by issues in scientific quality, and likely inability of current machine learning models to generalize to larger real-world datasets. Nevertheless, methods are steadily improving and it is important to ensure data privacy as much as possible before machine learning advances allow for disorder classification in reidentified EEG datasets. Efforts to support the protection of privacy include absolutely avoiding revealing direct personal reference, whether through deletion in the sense of de facto deletion or the activation of IT security measures. In addition, newer methods may allow the prevention of EEG data privacy breaches from the beginning, by removing or disguising identifying information from the EEG data to anonymize it before publishing (Debie et al., 2020; Liu et al., 2020; Schiliro et al., 2020; Popescu et al., 2021), or to only release necessary non-identifying features of the data upon request (Robinson & Varghese, 2016; Wang et al., 2022; Xia et al., 2022).

**Abbreviations**

ADHD: Attention deficit/hyperactivity disorder

ASD: Autism spectrum disorder

AUC: Area under the curve

AUPRC: Area under the precision recall curve

AUROC: Area under the receiver operating curve

CNN: Convolutional neural network

DA: Discriminant Analysis

EBT: Ensemble bagged tree

EEG: Electroencephalogram

EKNN: Ensemble k-nearest neighbor

GAD: Generalized anxiety disorder

PMP: In-phase matrix profile

HFD: Higuchi’s fractal dimension

iRBD: Idiopathic REM sleep disorder

IT: Information technology

KNN: K-nearest neighbor

LSTM: Long short-term memory

LR: Logistic regression

NB: Naïve Bayesian

NN: Neural network

MCI: Mild cognitive impairment

MCS: Minimally conscious state

MDD: Major depressive disorder

MLA: Modified logical accuracy

TBI:  Traumatic brain injury

OSAS: Obstructive sleep apnea

PCC: Pearson correlation coefficient

PI: Primary insomnia

REM: Rapid eye movement

REC: Resting eyes closed

ROC: receiver operating characteristics

REO: Resting eyes open

SAD: Social anxiety disorder

SVM: Support vector machine

TNN: Trilayer neural network

VS/UWS: Unresponsive wakefulness

**Supplementary Information**

Additional information about the quality analysis can be found in Supplementary_Material.pdf

## Declarations

**Ethics approval and consent to participate**

Not applicable. For individual studies this was assessed in the quality analysis (See Supplementary Material).

**Consent for publication**

Not applicable

**Availability of data and materials**

All data generated or analyzed during this study are included in this published article, and its supplementary information files.

**Competing interests**

The authors declare no competing interests, financial or otherwise, with any organization.

**Funding**

This work was funded by the German Federal Ministry of Education and Research and the European Union - NextGenerationEU, project NEMO, grant number 16KISA061K.

**Authors' contributions**

JEMS, AP, HG, TK, KCF, PA, and KIW all took part in the conceptualization of the review. JEMS curated the data, developed methodology, performed analysis, programmed visualizations and wrote the initial drafts of the paper. JEMS, MG, and AP performed the quality analysis. All authors took part in reviewing and editing the paper. KIW and PA acquired funding and supervised the project.

**Acknowledgements**

We would like to thank all partners of the NEMO (Nicht-Identifizierbarkeit von EEG-Daten und vergleichbaren Sensorsignalen aus medizinischer Versorgung für Open Science) project for support and discussions about this paper and topic.

**Table 1.** Summary of papers using resting state EEG to classify diseases and addictions.

| Study | Info classified | Sample size | Type of resting | Time (per subject) | Number of EEG sensors used | Best classification method & features | Best classification accuracy/results |
|---|---|---|---|---|---|---|---|
| **Liechti et al., 2013** | 54 ADHD vs. 51 Neurotypical | 62 children, 43 adults | EO & EC | 6 min (3 EO, 3 EC) | 1 | Discriminant analysis (DA) on spectral features | 53.2% |
| **Ciodaro et al., 2020** | 170 ADHD vs. 117 Neurotypical | 287 children | EC | 100 seconds | 110 | Extreme gradient boosting with polynomial delta and theta power band (ID-2) and visual representation of alpha and beta power with residuals CNN (convolutional neural network; ID-44) | ID-2: 86.3% precision 73.3% F1 score ID-44: 90% precision 76% F1 score |
| **Jennings et al., 2022** | 40 dementia patients (12 Alzheimer's disease, 21 dementia with Lewy bodies, 17 Parkinson's disease dementia), 15 healthy controls | 55 adults | EO & EC | 150s (75, 75) | 65 | NN, SVM and logistic regression classifier with spectral properties | Dementia vs. Healthy control: 91% |
| **Langer et al., 2022** | 1166 ADHD, 640 anxiety disorders, 439 learning disorders, 292 ASD, 297 disruptive disorders, 272 | 1921 children | EC (EO not used) | 200s, 100s | 105 | SVM, random forest (RF) & Neural networks classifiers with source and sensor level spectral features | AUPRC = 0.33 AUROC = 0.57 F1 score: 0.41 |
| | communication disorders, 194 depressive disorders, 542 other disorders, 206 with no diagnosis | | | | | | |
| **Wadhera, 2021** | 30 ASD vs. 30 Epilepsy, 30 ASD+Epilepsy, 30 typically developing | 120 children | Not stated | 1 min | 22 | SVM using average weighted degree, local efficiency, characteristic path length, and eigenvector centrality | 98.8% |
| **Park et al., 2021** | 117 schizophrenia, 128 trauma & stress related disorders, 199 anxiety disorders, 266 mood disorders, 186 addictive disorders, 46 OCD vs. 95 Healthy controls | 945 adults | EC | 5 min | 19 | Elastic net model using PSD and frequency band functional connectivity features | schizophrenia: 93.8%<br>trauma and stress-related disorders: 91.2%<br>anxiety disorders: 91.0%<br>mood disorders: 89.3%<br>addictive disorders: 85.7%<br>obsessive–compulsive disorder: 74.5% |
| **Garcés et al., 2022** | 212 ASD vs. 199 Neurotypical | 107 children, 145 adolescents, 159 adults | EO & EC | 4 min (2,2) | 61 | Elastic net logistic regression using power spectrum and functional connectivity features | 57% |
| **Ghoreishi et al., 2021** | 34 ASD vs. 27 Neurotypical | 61 children | EO | 1 min | 10 | SVM on Polar-based lagged state-space indices | 2 channel: 82%<br>1 channel: 78.7% |
| **Xia et al., 2023** | 30 MDD vs. 28 Healthy controls | 58 adults | EC | 5 min | 19 | End-to-end integrated deep learning model | 91.1% |
| **Rafiei et al., 2022** | 34 MDD vs. 30 Healthy controls | 64 adults | EC | 5 min | 19 | Deep convolutional neural networks and Choquet fuzzy integral on spectral features | 95.7% |
| **Uudeberg et al., 2024** | 33 MDD vs. 33 Healthy controls | 66 adults | EC | 6 min | 30 | In-phase matrix profile (pMP), Higuchi's fractal dimension (HFD) | PMp: 73%<br>HFD: 67% |
| **Wu et al., 2021** | 200 MDD vs. 200 Healthy controls | 400 adults | EO | 90 s | 26 | Conformal kernel-SVM using Band power, Higuchi's fractal dimension, coherence, and Katz's fractal dimension | Training set: 91.1%<br>Test set:<br>84.2% |
| **Sharma et al., 2021** | 16 MCI vs. 16 Dementia vs. 15 normals | 47 adults | EO & EC | 8 min (3, 3, 2) | 22 | SVM on PSD spectral features and fractal dimension using wavelet transform | MCI: 87.2% EO, 89.7% EC<br>Dementia: 88.7% EO, 90.2% EC<br>Normal: 94.7% EO, 90.2% EC |
| **Aljalal et al., 2024** | 32 MCI vs. 29 Healthy controls | 61 adults | EC | 10 min | 19 | Various classifiers including k-nearest neighbor (KNN) and ensemble KNN (ENKNN) with features extracted using variational mode decomposition (VMD) | KNN: 99.5%<br>ENKNN: 99.5% |
| **Kim et al., 2023** | Aβ+ vs. Aβ- pathology in 115 MCI (54 Aβ +, 61Aβ −) and 196 subjective cognitive decline (SCD)(36 Aβ +, 160 Aβ −), | 311 older adults | EC | 3 min | 19 | Various methods including SVM, Logistic, KNN, Random Forest, & feature selection models including T-test, ElasticNet, Whitney-Mann, Random Forest Importance, GBM, XGBoost with spectral features | 82.9% |
| **Shen et al., 2022** | 45 GAD, 36 healthy controls | 81 adults | EC | 10 min | 16 | SVM with Abberrated multidimensional EEG characteristic features | 97.8% |
| **Al-ezzi et al., 2021** | SAD (22 severe, 22 moderate, 22 mild) vs. 22 Healthy controls | 88 adults | EC | 6 min | 32 | CNN, LSTM, and CNN + LSTM using spectral effective connectivity features | Severe: 92.9%<br>Moderate: 92.9%<br>Mild: 96.4%<br>Healthy control: 89.3% |
| **Nour et al., 2023** | 15 Parkinson's disease vs. 16 Healthy controls | 31 adults | Not stated | 3 min | 32 | XGBoost model, Dynamic classifier selection in Modified Local Accuracy (MLA) and 1D-PDCovNN classification methods using common spatial pattern features | XGBoost: 98.7%<br>1D-PDCovNN: 98%<br>MLA: 99.3% |
| **Suuronen et al., 2023** | 89 EO + 68 EC Parkinson's patients, 89 EO + 63 EC control participants | 309 adults | EO & EC | 30s | 60 or less | Logistic regression classifier using varying channel numbers, complexity measures and spectral features | EO, 5 channel: 76%<br>EC, 56 channel: 72% |
| **Liu et al., 2017** | 17 Parkinson's patients and 25 healthy subjects | 42 adults | EC | 40s | 10 | optimal center constructive covering algorithm classifier with sample entropy features | 92.9% |
| **Kim et al., 2020** | 42 PTSD patients, 39 healthy individuals | 81 adults | EC | 3 min | 62 | Fischer geodesic minimum distance with source covariance features | 73.1% |
| **Khajehpour et al., 2019** | 36 meth-dependent individuals, 24 normals | 60 adults | EO | 5 min | 61 | SVM with graph theory and functional connectivity feature extraction | 93% |
| **Mumtaz et al., 2018** | 30 alcohol use disorder, 30 controls | 60 adults | EO & EC | 10 min (5,5) | 19 | Support Vector Machine (SVM), Naïve Bayesian (NB), and Logistic Regression (LR) with synchronization likelihood and spectral features | SVM: 98%<br>LR: 91.7%<br>NB: 93.6% |
| **Ahmadlou et al., 2013** | 36 methamphetamine abusers, 36 healthy subjects | 72 adults | EC | 180s | 32 | Enhanced Probabilistic Neural Network with functional connectivity and graph features | 82.8% |
| **Sharma & Acharya, 2021** | 14 schizophrenia, 14 normal | 28 adults | EC | 15 min | 1 | KNN with optimal wavelet-based l1 norm features | 99.2 % |
| **Kim et al., 2015** | 90 schizophrenia, 90 normal | 180 adults | EC | 3 min | 19 | Receiver operator characteristics (ROC) analysis with spectral features | Delta power: 62.2% |

**Table 2.** Summary of papers using sleep EEG to classify diseases and sleep disorders.

| Study | Info classified | Sample size | Sleep stages | Num. of EEG sensors | Time (per subject) | Classifier and features | Best Classification accuracy/results |
|---|---|---|---|---|---|---|---|
| **Aydin et al., 2015** | 7 psycho-physiological insomnia, 7 controls | 14 adults | Stage 1,2,3,4, Wake, REM, | 2 | 1 night sleep / 254-714 min | 10 classifiers using nearest neighbor methods, Bayesian methods, instance based methods, neural networks, regression methods, optimization methods with mutual information features, Welch method coherence features, Burg method coherence features, Pearson correlation coefficient (PCC) | Mutual information: 100%<br>Welch's coherence: 100%<br>Burg's coherence: 90%<br>PCC: 83% |
| **Sharma et al., 2023** | dataset 1: 9 insomnia, 6 healthy<br>dataset 2: 11 insomnia, 11 healthy | Dataset 1: 15 adults<br>Dataset 2: 22 adults | wake, N1, N2, N3, and REM | 1 | 9-10 hours | Ensemble bagged tree (EBT), weighted KNN, Trilayer neural network (TNN) with Wavelet Scattering Network features | Dataset 1: 96.5%<br>Dataset 2: 97% |
| **Qu et al., 2021** | Dataset 2: 11 insomnia, 11 healthy<br>Dataset 3: 30 insomnia, 18 healthy | Dataset 2: 22 adults<br>Dataset 3: 48 adults | Wake, N1, N2, N3, REM | 1 | 5 hours | LSTM classifier with feature extractor composed of feature extraction, feature transformation, and temporal context learning modules | Dataset 2: 90.9%<br>Dataset 3: 79.2% |
| **Yang & Liu, 2020** | 9 insomnia, 9 healthy subjects | 18 adults | Wake, S1, S2, S3, S4, REM | 1 | 7-14 hours | 1D-CNN based deep learning technique | REM: 87.5%<br>F1: 0.86 |
| **Hamida et al., 2016** | Training: 9 primary insomnia (PI), 8 control<br>Test: 10 PI, 8 controls<br>19 insomnia, 16 control subjects | 35 adults | Wake, S1, S2, SWS, REM | 1 | 1 night sleep | PCA classification Spectral & statistical features | SWS: 91% |
| **Kuo and Chen, 2020** | 16 insomnia, 16 control subjects (half training and half test) | 32 adults | Wake, N1, N2, N3, REM, and body movement | 1 (EOG) | < 30 mins | SVM with refined composite multiscale entropy analysis | 89% |
| **Shahin et al. (2017)** | 42 insomnia, 41 control participants | 83 adults | Wake, N1, N2, N3 or REM | 2 | 1 night sleep | DNN with statistical and spectral features | NREM+REM: 92% |
| **Wadichar et al., 2023** | 40 nocturnal frontal lobe | 108 adults | Cyclic alternating | 1 | Not specified | Long short-term memory (LSTM) with CNN | Phase B: |
| | epilepsy, 22 REM behaviour disorder, 10 periodic leg movement, 9 insomnia, 5 narcolepsy, 4 sleep-disordered breathing, 2 bruxism, 16 healthy controls | | pattern (CAP) phases A and B | | | | Healthy vs. Unhealthy: 92.7% Disease classification: 93.3 |
| **Malinowska et al., 2013** | 20 minimally conscious state (MCS), 11 vegetative/unresponsive wakefulness (VS/UWS) , 1 Locked-in Syndrome, 5 healthy controls | 32 adults | Not stated | 2 | > 9 hours | Discriminant analysis with variability and sleep related features | MCS vs. VS/UWS: 87% |
| **Moussa et al., 2022** | 40 obstructive sleep apnea (OSAS) and depression, 40 OSAS, | 118 adults | Light sleep, REM sleep, deep sleep | Full PSG ECG, 6 EEG, 6 breathing signals | 10-100 min | SVM classifier with various spectral and statistical features | Deep sleep: OSAS: 98.4% OSAS + depression: 73.8% |
| | 38 healthy controls | | | | | | |
| **Heyat et al., 2019** | 2 sleep bruxism, 6 healthy subjects | 8 adults | Stage 1, REM | 2 | 13-97 min | Decision tree with spectral features | S1: 85.8%<br>REM: 80.3%<br>Combined stages: 81.3% |
| **Bisgaard et al., 2015** | 34 idiopathic REM sleep Behaviour Disorder (iRBD), 42 control subjects | 76 adults | All, especially REM | 6 | 7-8 hours | K-Means clustering classifier with signal-to-noise correlation and peak amplitude features | 78% |
| **Hansen et al., 2013** | 10 iRBD patients, 10 control subjects | 20 adults | REM | 6 | 51.7-125.1 min | Bayesian classifiers with spectral features | 90% sensitivity<br>90% specificity |
| **Gholami et al., 2022** | 12 sleep apnea, 20 healthy subjects | 32 adults | Not stated | 3 | Not specified | SVM and KNN classifiers with minimum-redundancy maximum-relevance (mRMR) algorithm sorted complexity and entropy features | 93.3% |
| **Vishwanath et al., 2021** | 10 Traumatic brain injury (TBI) patients, 9 control subjects | 19 adults | Wake, N1, N2, N3, REM | 6 | 1 night sleep | RF, SVM and KNN classifiers Spectral features, Hjorth parameters and random sampling data arrangement | Random sampling (epochs): ~95%<br>Independent validation (subjects): ~70% |
| **De Dea et al., 2019** | 8 ADHD subjects and 9 healthy subjects (children) | 17 children | Sleep spindles, and time before and after | 19 | 40 min | Linear SVM with spectral features and principle component analysis (PCA) | 94.1% |
| **Melia et al., 2015** | 20 excessive daytime sleepiness patients, 20 patients without daytime sleepiness | 40 adults | Waking | 6 | 60s | Linear discriminant function using mutual information, and spectral features | Maintaining wakefulness (Channel O2):<br>0.83 AUC<br>65% sensitivity<br>75% specificity<br><br>Trying to sleep (O2):<br>0.94 AUC<br>80% sensitivity<br>90% specificity |

**Table 3.** Available databases with resting state and sleep EEG data.

| **Name** | **Reference** | **Permission needed?** | **Studies associated** |
|---|---|---|---|
| **Healthy brain network (HBN)** | (Healthy brain network, Alexander et al., 2017) | yes | Ciodaro et al., 2020, Langer et al., 2022; |
| **AI4Health** | (AI4Health) | no | Langer et al., 2022 |
| **MDD Patients and Healthy Controls EEG Data (New)** | (MDD Patients and Healthy Controls EEG Data (New)) | yes | Rafiei et al., 2022; Mumtaz et al., 2017 |
| **EEG Signals From Normal and MCI (Mild Cognitive Impairment) Cases** | (EEG Signals From Normal and MCI (Mild Cognitive Impairment) Cases; Kashefpoor et al., 2019) | no | Aljalal et al., 2024 |
| **UC San Diego Resting State EEG Data from Patients with Parkinson's Disease** | (UC San Diego Resting State EEG Data from Patients with Parkinson's Disease; Rockhill et al., 2021) | no | Nour et al., 2023 |
| **CAP sleep database** | (CAP sleep database; Goldberger et al., 2000; Terzano et al., 2001) | no | Yang & Liu, 2020; Sharma et al., 2023; Wadichar et al., 2023; Heyat et al., 2019 |
| **EEG/EOG/EMG data from a cross sectional study on psychophysiological insomnia and normal sleep subjects** | (EEG/EOG/EMG data from a cross sectional study on psychophysiological insomnia and normal sleep subjects; Rezaei et al., 2017) | no | Qu et al., 2021 |
| **Stanford Technology Analytics and Genomics in Sleep** | [Stanford Technology Analytics and Genomics in Sleep.] | yes | Zhang et al., 2018b; Moussa et al., 2022 |

Supplementary Material

For study assessment, we designed a custom assessment tool with 3 overall quality categories: high, medium, and low. Three reviewers independently performed study quality and publication bias assessments for each study, with two reviewers reviewing each paper and the third resolving conflicts between the other two.

*The following are multiple questions with **Yes**, **No**, or **Somewhat** as possible answers:*

1. Was the study objective or research question clearly stated?
2. Was the study population clearly defined? (e.g. age, sex, pathologies, recruiting ethics)
3. Did the study describe the classifier used?
4. Did the study report the amount of data used per subject for training and testing?
5. Did the study describe the disorder being classified?
6. Were the results/discussion well described & presented?
7. For machine learning papers, was the machine learning component well-executed? (e.g. description of epoch length, training/testing sets split appropriately, etc.)
8. Is the data available?
9. Is the code available?

To examine the potential of publication bias, we looked at each study's funding source, authors' conflicts of interest, and any other factors that could result in publication bias. We then rated each study into any of the three categories: ***Undetected, Suspected, Strongly Suspected.***

For the **overall quality,** we classified each paper into one of the three quality categories depending on the number of YES votes received from reviewers, resulting in the following grading scheme:

- High Quality: 6-9 YESs
- Moderate Quality: 3-5 YESs
- Low Quality: 0-2 YESs

Publication bias detection does not affect overall study quality unless it is strongly suspected, in which case the study loses 2 YES points. When answers were ‘not applicable’ (e.g. for question 7, some studies are not explicitly machine learning papers), this does not reduce a paper’s YES points.

| Resting Studies | 1 | 2 | 3 | 4 | 5 | 6 | 7 | 8 | 9 | Publication Bias | Overall Quality |
|---|---|---|---|---|---|---|---|---|---|---|---|
| Liechti et al., 2013 | 🟩 | 🟩 | 🟩 | 🟩 | 🟩 | 🟩 | ▨ | 🟥 | 🟥 | Undetected | High |
| Ciodaro et al., 2020 | 🟩 | 🟩 | 🟩 | 🟩 | 🟩 | 🟩 | 🟩 | 🟩 | 🟥 | Undetected | High |
| Jennings et al., 2022 | 🟩 | 🟩 | 🟩 | 🟩 | 🟩 | 🟩 | 🟨 | 🟨 | 🟨 | Undetected | High |
| Langer et al., 2022 | 🟩 | 🟩 | 🟩 | 🟩 | 🟩 | 🟩 | 🟩 | 🟩 | 🟩 | Undetected | High |
| Wadhera, 2021 | 🟩 | 🟨 | 🟩 | 🟩 | 🟩 | 🟩 | 🟩 | 🟥 | 🟥 | Undetected | High |
| Park et al., 2021 | 🟩 | 🟩 | 🟩 | 🟩 | 🟨 | 🟩 | 🟩 | 🟩 | 🟥 | Undetected | High |
| Garcés et al., 2022 | 🟩 | 🟩 | 🟩 | 🟩 | 🟩 | 🟩 | 🟩 | 🟨 | 🟥 | Suspected | High |
| Ghoreishi et al., 2021 | 🟩 | 🟩 | 🟩 | 🟩 | 🟩 | 🟩 | 🟩 | 🟥 | 🟥 | Undetected | High |
| Xia et al., 2023 | 🟩 | 🟩 | 🟩 | 🟩 | 🟩 | 🟩 | 🟩 | 🟩 | 🟥 | Undetected | High |
| Rafiei et al., 2022 | 🟩 | 🟩 | 🟩 | 🟩 | 🟩 | 🟩 | 🟩 | 🟩 | 🟥 | Undetected | High |
| Uudeberg et al., 2024 | 🟩 | 🟩 | 🟨 | 🟩 | 🟩 | 🟩 | 🟨 | 🟥 | 🟥 | Undetected | Moderate |
| Wu et al., 2021 | 🟩 | 🟩 | 🟩 | 🟩 | 🟩 | 🟩 | 🟩 | 🟥 | 🟥 | Undetected | High |
| Sharma et al., 2021 | 🟩 | 🟩 | 🟩 | 🟩 | 🟩 | 🟩 | 🟩 | 🟥 | 🟥 | Undetected | High |
| Aljalal et al., 2024 | 🟩 | 🟩 | 🟩 | 🟩 | 🟩 | 🟩 | 🟩 | 🟩 | 🟥 | Undetected | High |
| Kim et al., 2023 | 🟩 | 🟩 | 🟩 | 🟩 | 🟩 | 🟩 | 🟨 | 🟥 | 🟥 | Undetected | High |
| Shen et al., 2022 | 🟩 | 🟩 | 🟨 | 🟩 | 🟩 | 🟩 | 🟨 | 🟥 | 🟥 | Undetected | Moderate |
| Al-ezzi et al., 2021 | 🟩 | 🟩 | 🟩 | 🟩 | 🟩 | 🟩 | 🟨 | 🟨 | 🟥 | Undetected | High |
| Nour et al., 2023 | 🟩 | 🟩 | 🟩 | 🟩 | 🟩 | 🟩 | 🟨 | 🟩 | 🟥 | Undetected | High |
| Suuronenet al., 2023 | 🟩 | 🟩 | 🟨 | 🟩 | 🟩 | 🟩 | 🟩 | 🟥 | 🟥 | Undetected | High |
| Liu et al., 2017 | 🟩 | 🟩 | 🟩 | 🟩 | 🟩 | 🟩 | 🟩 | 🟥 | 🟥 | Undetected | High |
| Kim et al., 2020 | 🟩 | 🟩 | 🟩 | 🟩 | 🟩 | 🟩 | 🟩 | 🟥 | 🟩 | Undetected | High |
| Khajehpour et al., 2019 | 🟩 | 🟩 | 🟩 | 🟩 | 🟩 | 🟩 | 🟨 | 🟥 | 🟥 | Undetected | High |
| Mumtaz et al., 2018 | 🟩 | 🟩 | 🟩 | 🟩 | 🟩 | 🟩 | 🟩 | 🟥 | 🟨 | Undetected | High |
| Ahmadlou et al., 2013 | 🟩 | 🟩 | 🟨 | 🟩 | 🟩 | 🟩 | 🟩 | 🟥 | 🟥 | Undetected | High |
| Sharma & Acharya, 2021 | 🟩 | 🟨 | 🟩 | 🟩 | 🟩 | 🟩 | 🟩 | 🟥 | 🟥 | Undetected | High |
| Kim et al., 2015 | 🟨 | 🟩 | 🟨 | 🟩 | 🟩 | 🟨 | 🟥 | 🟥 | 🟥 | Undetected | Moderate |

**Supplementary Figure 1.** Paper quality and publication bias assessments for studies using resting state EEG data. Green squares indicate a ’yes’ to the respective question, yellow indicates ’somewhat’, red indicates ’no’, and grey indicates ’not applicable’.

| Sleep Studies | 1 | 2 | 3 | 4 | 5 | 6 | 7 | 8 | 9 | Publication Bias | Overall Quality |
|---|---|---|---|---|---|---|---|---|---|---|---|
| Aydin et al., 2015 | | | | | | | | | | Undetected | Low |
| Sharma et al., 2023 | | | | | | | | | | Undetected | High |
| Qu et al., 2021 | | | | | | | | | | Undetected | High |
| Yang & Liu, 2020 | | | | | | | | | | Undetected | High |
| Hamida et al., 2016 | | | | | | | | | | Undetected | Moderate |
| Kuo and Chen, 2020 | | | | | | | | | | Undetected | Moderate |
| Shahin et al., 2017 | | | | | | | | | | Undetected | Moderate |
| Wadichar et al., 2023 | | | | | | | | | | Undetected | High |
| Malinowska et al., 2013 | | | | | | | | | | Undetected | High |
| Moussa et al., 2022 | | | | | | | | | | Undetected | High |
| Heyat et al., 2019 | | | | | | | | | | Undetected | Moderate |
| Bisgaard et al., 2015 | | | | | | | | | | Undetected | Moderate |
| Hansen et al., 2013 | | | | | | | | | | Undetected | Moderate |
| Gholami et al., 2022 | | | | | | | | | | Undetected | Moderate |
| Vishwanath et al., 2021 | | | | | | | | | | Undetected | Moderate |
| De Dea et al., 2019 | | | | | | | | | | Undetected | Moderate |
| Melia et al., 2015 | | | | | | | | | | Undetected | High |

**Supplementary Figure 2.** Paper quality and publication bias assessments for studies using sleep EEG data. Green squares indicate a ’yes’ to the respective question, yellow indicates ’somewhat’, red indicates ’no’, and grey indicates ’not applicable’.